\documentclass[10pt,twocolumn,letterpaper]{article}

\usepackage[pagenumbers]{wacv} 

\usepackage{graphicx}
\usepackage{algorithm}
\usepackage{algorithmic}
\usepackage{xcolor}     
\usepackage[table]{xcolor}
\usepackage{multirow}
\definecolor{wacvblue}{rgb}{0.21,0.49,0.74}
\usepackage[pagebackref,breaklinks,colorlinks,allcolors=wacvblue]{hyperref}

\def\wacvPaperID{2949} 
\def\confName{WACV}
\def\confYear{2027}

\title{ECOKV: Geometry-Aware KV Cache Eviction via Complementary Diversity Metrics}

\author{%
  \textbf{Chin Ting Hsu}$^{1}$ \quad \quad \quad
  \textbf{Yu-Syuan Xu}$^{1,2}$ \quad \quad \quad
  \textbf{Ling Zou}$^{1}$ \\
  \textbf{Hsien-Kai Kuo}$^{2}$ \quad \quad \quad
  \textbf{Wen-Huang Cheng}$^{1}$ \\ [10pt]
  $^{1}$National Taiwan University \quad
  $^{2}$MediaTek Inc.\quad
}

\begin{document}
\maketitle
\begin{abstract}
Although multimodal Large Language Models (MLLMs) excel in diverse tasks, their scalability remains limited by the memory and computational overhead of KV cache storage. Recent KV cache eviction approaches incorporate a cosine similarity-based diversity metric with importance metrics to selectively retain critical key-value pairs. However, cosine similarity involves normalization that discards magnitude information, and it often yields uniformly high similarity values across layers due to the anisotropy property of hidden representations. In our study ECOKV, we rigorously deconstruct the capabilities of existing diversity metrics. Moving beyond simple measurement, we propose a geometry-aware composite metric that jointly leverages Euclidean distance and cosine similarity to capture token diversity from complementary perspectives. Furthermore, we use these two metrics to estimate the redundancy level of each attention head, allowing adaptive weighting between diversity and importance scores during token selection. Finally, we demonstrate that the observation window commonly employed to preserve recent tokens can be substantially reduced, thereby allocating more cache capacity to informative tokens and yielding consistent improvements. Extensive experiments demonstrate that ECOKV achieves state-of-the-art performance under various compression ratios and can be seamlessly integrated with existing KV cache eviction methods. We further analyze the relationship between importance and diversity, and examine redundancy patterns across layers and attention heads.
\end{abstract}
    
\section{Multimodal Large Language Model Efficiency}

Multimodal large language models (MLLMs)~\citep{Yang2023TheDO, zhang2024mm, yin2024survey} have gained widespread attention for their ability to process and reason over multimodal inputs, including text, images, and audio, while providing contextually relevant responses. In the realm of computer vision, they have been widely adopted in various tasks, including visual question answering~\citep{antol2015vqa, 10.1145/3711680}, image captioning~\citep{ghandi2023deep, hossain2019comprehensive}, and object detection~\citep{kaur2023comprehensive, zou2023object}. Recent advances have enabled the processing of increasingly long and complex inputs~\citep{Song2024MileBenchBM, bai2025qwen25vltechnicalreport}. However, this progress introduces substantial computational and memory overhead, making inference efficiency a critical concern. To address this challenge, KV cache eviction methods~\citep{jiang2026towards,li2024survey} offer a training-free alternative that reduces memory usage and inference latency while preserving model capability, by selectively removing redundant key-value pairs from the KV cache. 

Early KV cache eviction algorithms are primarily designed for Large Language Models (LLMs). They perform top-K selection within a fixed budget, retaining key-value pairs by importance scores estimated via accumulated attention score~\citep{zhang2023h2o}, average attention scores within an observation window~\citep{li2024snapkv}, or approximated by heuristic metrics such as Key Norm (KNorm)~\citep{devoto-etal-2024-simple} or Value Norm (VNorm)~\citep{kim2025infinipotv}. To account for the varying information density across layers and attention heads, subsequent work explores dynamic budget allocation frameworks that support adaptive and fine-grained budget sharing~\citep{feng2024ada, cai2024pyramidkv}. Recently, diversity-aware criteria have been incorporated to improve coverage of the retained key-value set and mitigate redundancy~\citep{park2025keydiff, datta2026manifoldkv}, yielding better preservation of the token distribution under constrained memory budgets. Recent studies have begun exploring KV cache eviction for MLLMs~\citep{wan2024look, liu2026mixing, li2025madakv, wang2025sparsemm, wan2025meda}. These works observe that visual tokens often exhibit higher redundancy than text tokens, motivating strategies that prioritize the preservation of text tokens or leverage diversity metrics to identify and remove redundant visual tokens. \\

\begin{figure}[t]
  \centering
  
  \includegraphics[width=0.9\linewidth]{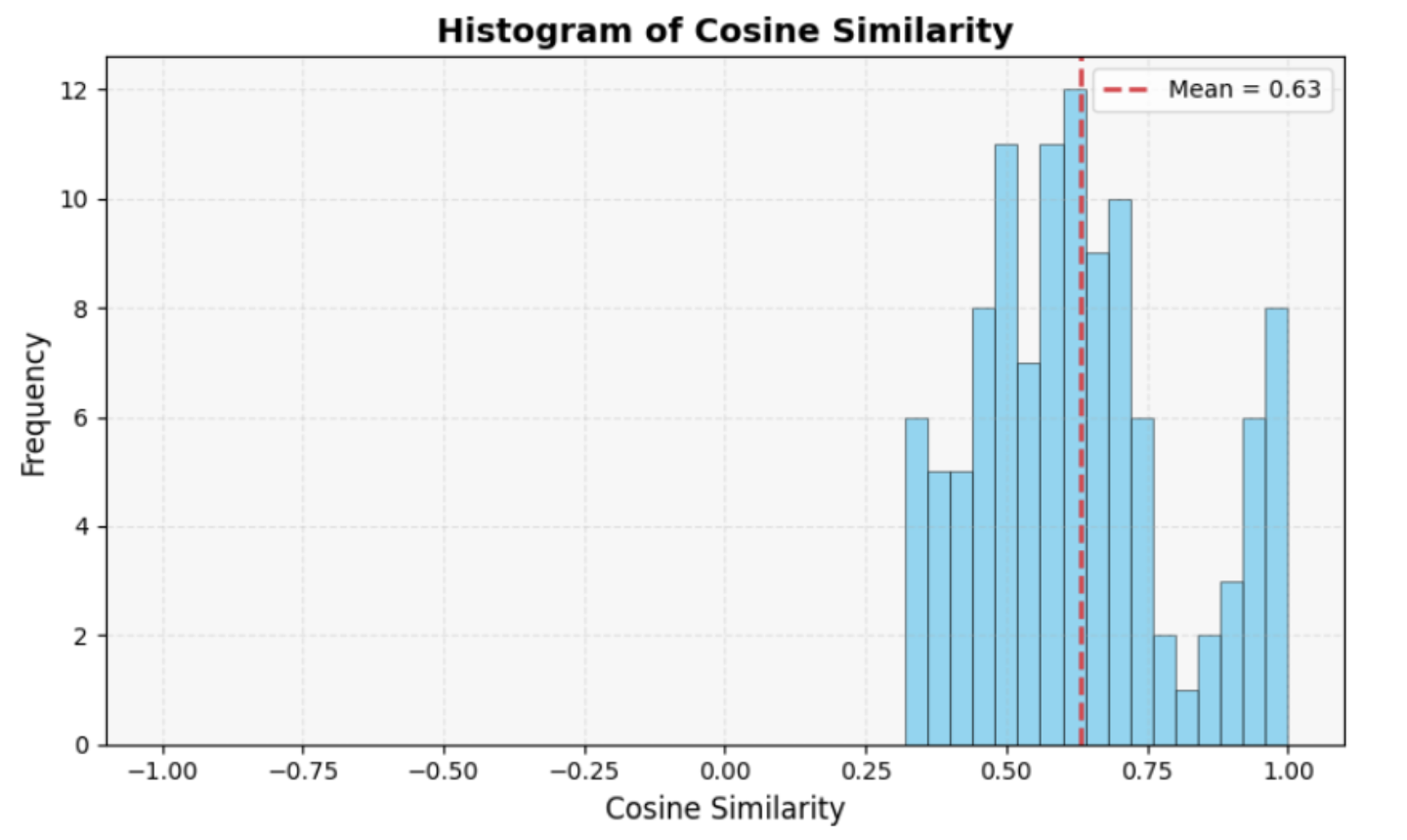}
  \caption{Distribution of layer-wise key cosine similarities in Qwen2-VL-7B-Instruct. Cosine similarity remains consistently high across all layers.
}
  \label{fig:teaser}

\end{figure}

Despite these advances, existing diversity-based methods mostly rely on cosine similarity over normalized embeddings, which introduces a fundamental geometric limitation. Recent studies~\citep{liang2022mind, godey2024anisotropy} have highlighted the issue of anisotropy, where high-dimensional transformer embeddings concentrate within a narrow cone, resulting in uniformly high cosine similarity scores. As illustrated in Figure~\ref{fig:teaser}, cosine similarity produces consistently high average similarity scores across layers, exposing its tendency to overestimate redundancy and potentially corrupt eviction decisions. Moreover, cosine similarity operates on normalized vectors, thereby losing magnitude information. Relying on a single, biased metric without jointly accounting for complementary geometric information can lead to incorrect token prioritization during eviction, while overestimation of redundancy may induce an undue preference for diversity over importance across layers. Euclidean distance offers a potential measure of token diversity by capturing magnitude information in token representations. However, lacking explicit directional information, it has shown inferior performance for token diversity assessment in prior LLM studies~\citep{park2025keydiff}. Consequently, each metric captures distinct information while having inherent limitations, and the potential benefits of combining them into a composite metric remain largely unexplored.

To address the limitations, we propose Geometry-Aware KV Cache Eviction via Complementary Diversity Metrics (ECOKV), a novel eviction framework that integrates Euclidean distance and cosine similarity to capture richer geometric information for token diversity. Specifically, we introduce a Euclidean-based diversity metric that measures the distance between each key and the global average, followed by normalization to facilitate its integration with a cosine-similarity-based metric. Furthermore, we employ Euclidean distance to capture layer-wise redundancy while mitigating its unbounded scale issue. In addition, previous methods inherit the observation window size of 32 from \citep{li2024snapkv}, which has not been thoroughly examined for MLLMs under tight memory budgets. With a small cache budget, the observation window consumes a substantial portion of the cache, leaving fewer tokens for selection. We therefore evaluate different observation window sizes and find that smaller windows consistently improve performance by allocating more budget to our token selection strategy.

Extensive experiments validate that ECOKV achieves state-of-the-art performance across multiple multimodal benchmarks and compression ratios, integrating seamlessly with existing eviction frameworks. We further analyze the relative weighting between Euclidean distance and cosine similarity, the effect of observation window size, and the layer-wise correlation between importance and diversity, offering insights into where diversity plays a critical role in key-value selection. Our key contributions are: 

\begin{itemize}
    \item \textbf{(i)} We provide in-depth analysis on different diversity metrics and their impact on model performance
    \item \textbf{(ii)} We introduce a head-adaptive key diversity and redundancy estimator that jointly leverages Euclidean distance and cosine similarity to dynamically adjust diversity contribution across attention heads
    
    \item \textbf{(iii)} We observe that a reduced observation window strategy can improve performance under tight budget constraints
    
    \item \textbf{(iv)} We achieve state-of-the-art performance on multimodel question answering benchmarks, and provide diverse empirical analysis on the choice of diversity implementation and the relation between different metrics
    
\end{itemize}

\section{Related Work}

\subsection{LLM-based KV Cache Eviction}

Recent work has explored KV cache eviction to reduce the memory overhead of long-context inference in LLMs~\citep{li2024survey, zhou2024dynamickv, feng2025identify, 10.5555/3692070.3694025, liu2024minicache, adnan2024keyformer}. H2O~\citep{zhang2023h2o} identifies a small subset of tokens that dominate attention and prioritizes them as heavy hitters. SnapKV~\citep{li2024snapkv} shows that only a subset of prompt tokens carries essential information for generation and compresses the KV cache by retaining representative tokens. They assess the importance of keys and values using the observation window located at the end of the prompts. Subsequent works adopt these ideas and explore dynamic budget allocation across attention layers and heads.  PyramidKV~\citep{cai2024pyramidkv} reveals that the attention is diffuse in the early layers and gradually concentrates in the deep layers, thereby allocating more cache in the lower layers and less in the higher ones. AdaKV~\citep{feng2024ada} finds that attention heads exhibit diverse concentration patterns and dynamically allocates cache budgets across heads accordingly. Another research direction investigates the estimation of token importance to determine which key-value pairs should be retained. InfiniPot-V~\citep{kim2025infinipotv} uses VNorm (value norm) as an importance proxy and introduces adaptive pooling to model spatial locality patterns across layers. Alternatively, \citep{devoto-etal-2024-simple} proposes KNorm (key norm) as the importance indicator, demonstrating that different components of KV representations can serve as effective signals for identifying important tokens.

For diversity-based assessment, KeyDiff~\citep{park2025keydiff} proposes an attention-free cache eviction method based on key similarity. R-KV~\citep{cai2026r} balances redundancy and importance with a hyperparameter $\lambda$ and performs eviction during decoding. ManifoldKV~\citep{datta2026manifoldkv} ranks tokens solely based on their Euclidean distance to the key centroid, overlooking both token importance and directional information such as cosine similarity. Although they achieve decent performance in LLMs, directly applying them to MLLMs can lead to a performance drop due to distinct modality-specific characteristics. \\

\subsection{MLLM-based KV Cache Eviction}

For MLLMs KV cache eviction, recent work aims to recognize the different characteristics between text and vision tokens. LOOK-M~\citep{wan2024look} observes that models often prioritize more textual attention over image features, inspiring them to prioritize text KV pairs during eviction. MEDA~\citep{wan2025meda} extends the idea by introducing a dynamic layer-wise KV cache allocation mechanism. SparseMM~\citep{wang2025sparsemm} uses an anchor task to identify visual heads and allocates them a larger budget, as they are primarily responsible for processing visual information. SimiCache~\citep{liu2026simicache} first applies K-means clustering, followed by similarity-based clustering to group similar keys, and then merges them through a weighted sum. MadaKV~\citep{li-etal-2025-madakv} allocates separate budgets for different modalities and dynamically adjusts the budget size based on modality complexity across layers. MixKV~\citep{liu2026mixing} mixes importance with diversity and dynamically determines the weight of each metric. These methods acknowledge the high redundancy of vision tokens, but either overlook diversity or rely on incomplete geometric information, leading to inaccurate diversity estimation. 

\subsection{Similarity Metrics in Transformers}
Similarities between tokens or hidden embeddings are frequently used in LLMs and MLLMs research~\citep{liu2026mixing, park2025keydiff, datta2026manifoldkv}. Cosine similarity is believed to capture semantic-level similarity and is therefore commonly used in NLP research~\citep{reimers2019sentence,mahajan2025revisiting}. However, it sometimes performs worse than other approaches~\citep{karpukhin2020dense, khattab2020colbert}. \citep{steck2024cosine} proposes that cosine similarity is implicitly controlled by regularization when learning deep models. In fact, cosine similarity in transformers is inherently high across various settings, a phenomenon known as anisotropy~\citep{liang2022mind, godey2024anisotropy}. In addition to cosine similarity, Euclidean distance is also adopted in some papers. For example, \citep{tessari2024surpassing} shows that using a distance-based Euclidean metric yields a similarity measure less biased by the dimensionality of the hidden-state. For MLLMs KV Cache eviction, recent research~\citep{liu2026mixing, liu2026simicache} inherits the assumption from LLMs and continues to use cosine similarity as the primary metric. In our paper, we propose using both Euclidean and cosine similarity metrics, which complement each other, to demonstrate that both are important for assessing token diversity in MLLMs.
\section{Preliminary}

Inference in LLMs and MLLMs can be divided into two stages: the prefilling stage and the decoding stage. The prefilling stage processes the entire input prompt, while the decoding stage autoregressively generates new tokens.
During the prefilling stage, for each transformer layer, the input $x \in \mathbb{R}^{T \times d}$ is projected through learnable weight matrices $W_q \in \mathbb{R}^{d \times d_q}$, $W_k \in \mathbb{R}^{d \times d_k}$, and $W_v \in \mathbb{R}^{d \times d_v}$ to obtain the query $Q$, key $K$, and value $V$ representations for self-attention which is computed as: \\
\begin{equation}
\begin{aligned}
   Q = x \cdot W_q, \quad K = x \cdot W_k, \quad V = x \cdot W_v, \\
   \text{Attention}(Q, K, V) = \text{softmax}\!\left(\frac{QK^\top}{\sqrt{d_k}}\right)V  \nonumber
\end{aligned}
\end{equation} 
\\
To avoid recomputing $K$ and $V$ at each decoding step, the KV cache retains these representations across the entire generation process. However, as input context lengths grow, the KV cache size increases linearly with the sequence length, imposing substantial memory overhead and computational cost. To mitigate this, KV cache eviction methods selectively remove less informative key-value pairs to maintain a bounded cache size while preserving model performance.

Let $K^{l,h} = \{K^{l,h}_i\}_{i=1}^{T}$ and $V^{l,h} = \{V^{l,h}_i\}_{i=1}^{T}$ denote the keys and values at layer $l$ and attention head $h$, respectively, where $T$ denotes the sequence length. We define a scoring function $S(K^{l,h}_i, V^{l,h}_i)$ that assigns a scalar score to each key-value pair, retaining only the top-K pairs with the highest scores. The selected subset is defined as: $ \hat{K}^{l,h}, \hat{V}^{l,h} = \{ K^{l,h}_i, V^{l,h}_i \;|\; S  \; (K^{l,h}_i, V^{l,h}_i) \in \text{Top-k} \}_{i=1}^{T}$, which is subsequently used to compute self-attention.

The scoring function $S$ typically quantifies either importance or diversity. Importance-based scoring employs an observation window of the most recent $N$ key-value pairs to assess the importance of remaining pairs, preserving the window unconditionally due to recency and relevance. Diversity-based scoring assigns higher scores to outlier tokens to promote semantic diversity in the retained set. Combining both criteria, the final set of preserved key-value pairs is given by: \\
\begin{flalign} 
\hat{K}^{l,h}, \hat{V}^{l,h} &= \underbrace{ \left\{ K^{l,h}_i, V^{l,h}_i \;\middle|\; S\left(K^{l,h}_i, V^{l,h}_i\right) \in \text{Top-}k \right\}_{i=1}^{T-N} }_{\text{diversity/importance-selected pairs}} \nonumber \\
& \cup\; \underbrace{ \left\{ K^{l,h}_i, V^{l,h}_i \right\}_{i=T-N+1}^{T} }_{\text{observation window}} \nonumber
\end{flalign} 
\\
where the first term contains the top-$k$ selected pairs from the non-window tokens, and the second term corresponds to the observation window that is preserved in its entirety. 




\section{Diversity Metrics Analysis} \label{observation}

Diversity metrics are widely adopted in token reduction and KV cache eviction methods to minimize information loss and mitigate redundancy within the retained subset. In the following paragraphs, we systematically analyze several commonly employed metrics for assessing token-level diversity, highlighting their respective strengths and limitations.



\paragraph{Cosine Similarity.} Cosine similarity measures the angular alignment between two vectors, making it sensitive to directional differences while remaining invariant to vector magnitudes. Given two vectors $a,b$, it is defined as:

\[
\text{CosSim}(\mathbf{a}, \mathbf{b}) = \frac{\mathbf{a} \cdot \mathbf{b}} {\|\mathbf{a}\| \|\mathbf{b}\|} 
\] 

It is widely adopted in NLP research and is often regarded as a proxy for semantic similarity between text tokens. However, by operating on normalized vectors, cosine similarity discards magnitude information, providing only a partial geometric information of token diversity. 

\begin{figure}[t]
  \centering
  \includegraphics[width=\linewidth]{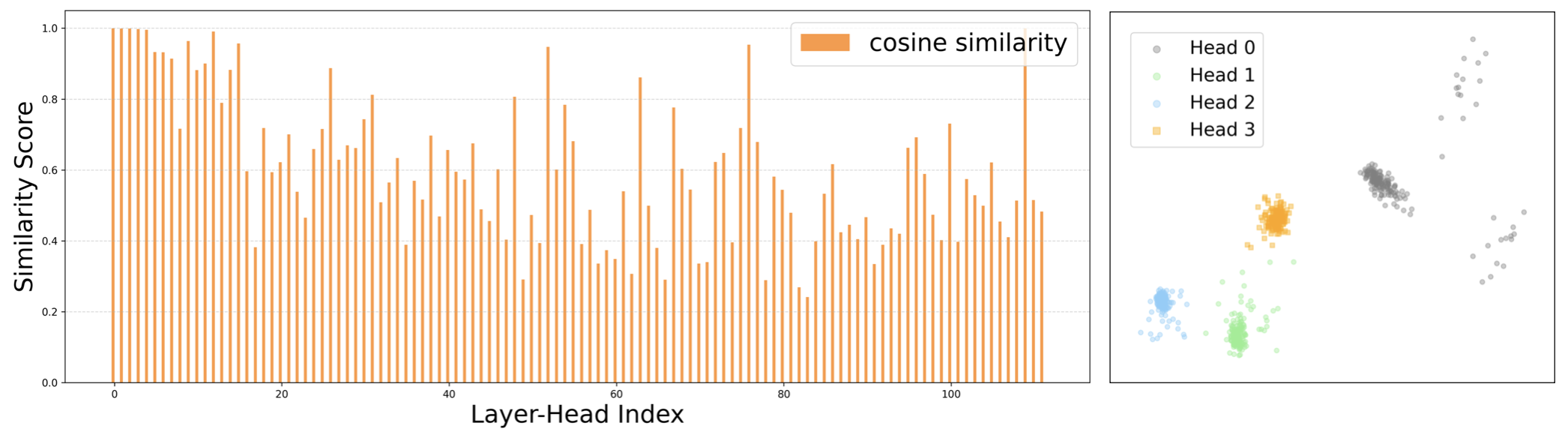}
  \caption{(a) Cosine similarity remains consistently high across layers. (b) PCA projection of layer 0 keys. Although the embeddings are not identical, their cosine is approximately 1.0.}
    \label{figure:cosine_problem}
\end{figure}

Moreover, recent work~\citep{liang2022mind,godey2024anisotropy,tessari2024surpassing} shows that high-dimensional embeddings in transformers tend to concentrate within a narrow cone, leading to consistently high average cosine similarity. Figure \ref{figure:cosine_problem} illustrates the cosine similarity across layers in Qwen-7B-Instruct~\citep{wang2024qwen2}. We observe that the average similarity exceeds $0.4$ in all layers and approaches $1.0$ in the first layer. In contrast, PCA visualizations of the first layers reveal that the token representations are not identical but remain well spread in space. This discrepancy highlights a key limitation: by ignoring magnitude information, cosine similarity can overestimate the similarity between embeddings.


\paragraph{Euclidean Distance.} Euclidean distance measures the straight-line distance between two vectors, capturing differences in both magnitude and position: 
\begin{equation} 
\text{ECD}(\mathbf{a}, \mathbf{b}) = \|\mathbf{a} - \mathbf{b}\|_2 = \sqrt{(\mathbf{a} - \mathbf{b})^\top(\mathbf{a} - \mathbf{b})}  \nonumber
\end{equation} 
Unlike cosine similarity, Euclidean distance operates on unnormalized vectors, thereby preserving magnitude information. This relationship can be made explicit by expanding the squared distance: 
\begin{equation} 
\text{ECD}(\mathbf{a}, \mathbf{b})^2 = \|\mathbf{a}\|^2 + \|\mathbf{b}\|^2 - 2\|\mathbf{a}\|\|\mathbf{b}\|\,\text{CosSim}(\mathbf{a}, \mathbf{b}) \nonumber
\end{equation}
This decomposition reveals that Euclidean distance jointly encodes vector magnitudes ($\|\mathbf{a}\|^2$, $\|\mathbf{b}\|^2$) and directional relationship, capturing richer geometric information than cosine similarity alone. Nonetheless, Euclidean distance is unbounded, which necessitates an additional mapping from distance to a normalized similarity measure. Furthermore, its scale can vary significantly across layers, making direct comparison of diversity between layers challenging.

\section{Method}

In this section, we propose a novel combined metric for MLLM KV Cache Compression by addressing two fundamental challenges: (a) how to design diversity metrics that assign reliable scores to each key (Sec.~\ref{subsec:EDD}), and (b) how to assess layer-wise redundancy and appropriately balance diversity and importance scores (Sec.~\ref{subsec:EDHR}). Furthermore, we introduce our findings regarding observation window size adjustment in Sec.~\ref{subsec:LDWC}.

\subsection{Key Diversity Score Assessment} 
\label{subsec:EDD} 

As analyzed in the previous section, cosine similarity suffers from anisotropy and lacks magnitude information. To provide enriched geometry information for key diversity score assessment, we incorporate Euclidean distance-based diversity metric as a complementary counterpart to cosine similarity. Specifically, we compute the centroid $\bar{k}$ of all key vectors and measure the Euclidean distance of each key $k_i$ from the centroid: 

\begin{equation} 
\bar{k} = \frac{1}{T} \sum_{i=1}^{T} k_i, \quad d_E(i) = \left\lVert k_i - \bar{k} \right\rVert_2 \\
\end{equation}

where $d_E(i)$ denotes the Euclidean distance between key $k_i$ and the centroid $\bar{k}$. A larger $d_E(i)$ indicates that $k_i$ deviates more from the average representation, suggesting it carries more unique and non-redundant information.

To eliminate the scale bias of Euclidean distance and map the distances to diversity scores from $0$ to $1$, we apply min-max normalization with a small constant $\epsilon$ for numerical stability: 
\begin{equation} 
s_E(i) = \frac{d_E(i) - \min(d_E)}{\max(d_E) - \min(d_E) + \epsilon} 
\end{equation}

A higher $s_E(i)$ corresponds to a key that deviates more from the centroid, indicating a higher diversity contribution. Normalization will not change the relative order of the tokens, but is crucial for the following combination with cosine similarity-based scores.

We then combine $s_E(i)$ with the cosine similarity-based diversity score $s_C(i)$ from \citep{liu2026mixing} (details provided in supplementary section \ref{baselines}), forming a unified diversity score that jointly captures both magnitude- and direction-aware information: 
\begin{equation} S_{div}(i) = \alpha \cdot s_E(i) + (1 - \alpha) \cdot s_C(i) \end{equation}
where $\alpha \in [0, 1]$ is a combination coefficient that balances the contribution of magnitude-aware Euclidean diversity and direction-aware cosine diversity. $S_{div}(i)$ is the diversity score for the $i_{th}$ key value pairs and will be further combined with the important score to derive the final selection priority.\\

\subsection{Head-adaptive Redundancy Estimation} 
\label{subsec:EDHR} 

Next, we combine diversity-based scores with importance scores. Intuitively, layers or attention heads with higher redundancy should place greater emphasis on diversity to reduce redundant KV pairs. However, high cosine similarity does not necessarily indicate high redundancy, as it is an inherent property of transformer representations. Therefore, we design separate heuristics that combine different metrics to better approximate semantic-level diversity. We present the most effective approach here and defer alternative designs to supplementary section \ref{alter}.

We propose a head-adaptive redundancy estimator that quantifies per-head redundancy by jointly leveraging Euclidean distance and cosine similarity to adaptively determine the diversity weight $\lambda$ for each attention head.

  For each attention head $h$, we compute the mean Euclidean distance $D_E^h$ across all key vectors as a measure of the overall spread of the key distribution: 
\begin{equation} 
D_E^h = \frac{1}{T} \sum_{i=1}^{T} d_E^h(i) 
\end{equation} 
where a larger $D_E^h$ indicates lower redundancy and a smaller $D_E^h$ implies higher redundancy. $D_E^h$ is then mapped to a normalized magnitude-aware redundancy score $r_E \in (0, 1)$ via a scaled sigmoid: 
\begin{equation} 
r_E = 2 - 2 \cdot \sigma\!\left(\frac{D_E^h}{\tau}\right)
\end{equation} 
where $\tau$ is a temperature parameter controlling estimation sensitivity. We choose $\tau=\sqrt{d_k}$, as the average distance scales proportionally with the square root of the key dimension. $r_E$ is then combined with the cosine similarity-based redundancy score $r_C$ from \citep{liu2026mixing} (details provided in supplementary section \ref{baselines}) to form a unified redundancy estimate that jointly captures magnitude and directional information: 
\begin{equation} 
\lambda = \beta \cdot r_E + (1 - \beta) \cdot r_C 
\end{equation} 
where $\beta \in [0, 1]$ balances magnitude-aware and direction-aware contributions. More redundant heads receive a larger $\lambda$ to emphasize diversity-based selection, and less redundant heads receive a smaller $\lambda$ to prioritize importance-based scoring. 

The complete algorithm for deriving $S$ and $\lambda$ is provided in supplementary section \ref{alg}, and sensitivity analyses of $\alpha$ and $\beta$ are presented in Sec.~\ref{subsec:ablation}.

Before combination, the diversity score $S_{div}$ is further scaled to match the magnitude of importance scores.

    \[
    \bar{S_{div}} = \frac{S_{div} - min}{max-min}, \quad \tilde{S_{div}} = \bar{S_{div}} \cdot \frac{mean(S_{imp})}{mean(\bar{s_C})}
    \]
where $S_{\mathrm{imp}}$ denotes the importance score, computed by combining the $\ell_2$ norm of the value vector with the average attention score estimated over the observation window, following previous works. The final score combines the scaled diversity score $\tilde{S}_{div}(i)$ and importance score $S_{imp}(i)$  with the head-adaptive weight $\lambda$:
\begin{equation} 
S(i) = \lambda \cdot \tilde{S}_{div}(i) + (1 - \lambda) \cdot S_{imp}(i) 
\end{equation} 

The top-$k$ pairs with the highest $S(i)$ are retained for each attention head, enabling fine-grained adaptive eviction across layers and heads.

\subsection{Minimized Observation Window}
\label{subsec:LDWC}

Previous work utilizes an observation window to assess the importance of each token. The token within the observation window is preserved, assuming that recent tokens are inherently more important. However, across different benchmarks and modalities, the last few tokens are not always the most significant. Instead, arbitrarily keeping these tokens can reduce the budget for dynamic token selection.

Our experiments show that reducing the observation window to a very small size—or even preserving only the most recent token—can lead to substantially better performance. To investigate this phenomenon, we conduct a detailed analysis of model performance across varying observation window sizes and KV budgets, as well as token selection distributions across benchmarks. The results demonstrate that a tiny observation window enables the KV cache eviction strategy to make more flexible and effective decisions about which tokens to retain, thereby improving utilization of the limited KV budget.

\begin{table*}[h]
\centering
\caption{Performance comparison across benchmarks under different methods and models.}
\setlength{\tabcolsep}{3pt}
\scalebox{0.93}{
\begin{tabular}{l|ccc|ccc|ccc|ccc|ccc}
\toprule
\textbf{Methods} 
& \multicolumn{3}{c}{\textbf{DocVQA (\%)}} 
& \multicolumn{3}{c}{\textbf{OCRBench (\%)}} 
& \multicolumn{3}{c}{\textbf{TextVQA (\%)}} 
& \multicolumn{3}{c}{\textbf{ChartQA (\%)}} 
& \multicolumn{3}{c}{\textbf{TextCaps}} \\
\cmidrule(lr){2-4} \cmidrule(lr){5-7} \cmidrule(lr){8-10} \cmidrule(lr){11-13} \cmidrule(lr){14-16}
& 256 & 128 & 64 & 256 & 128 & 64 & 256 & 128 & 64 & 256 & 128 & 64 & 256 & 128 & 64 \\
\midrule

\multicolumn{16}{c}{\textbf{LLaVA-NeXT-Mistral-7B}} \\
\midrule
SnapKV 
& 59.7 & 55.2 & 47.3 & 45.0 & 39.0 & 31.9 & 63.5 & 61.0 & 57.1 & 50.2 & 47.5 & 42.7 & 0.650 & 0.558 & 0.444 \\
+ MixKV 
& 61.7 & 58.1 & 48.8 & \textbf{49.9} & 44.7 & 36.1 & 65.2 & 64.3 & 60.1 & 50.8 & 47.7 & 43.6 & 0.708 & 0.659 & 0.514 \\
\rowcolor{gray!10}
+ ECOKV
& \textbf{62.7} & \textbf{60.8} & \textbf{56.9} & 49.7 & \textbf{48.4} & \textbf{43.7} & \textbf{65.5} & \textbf{65.1} & \textbf{63.0} & \textbf{52.5} & \textbf{51.5} & \textbf{48.8} & \textbf{0.792} & \textbf{0.721} & \textbf{0.678} \\
\rowcolor{gray!20}
$\Delta_{\text{baseline}}$
& \textcolor{green!60!black}{+1.0} 
& \textcolor{green!60!black}{+2.7} 
& \textcolor{green!60!black}{+8.1} 
& \textcolor{red!60!black}{-0.2} 
& \textcolor{green!60!black}{+3.7} 
& \textcolor{green!60!black}{+7.6} 
& \textcolor{green!60!black}{+0.3} 
& \textcolor{green!60!black}{+0.8} 
& \textcolor{green!60!black}{+2.9} 
& \textcolor{green!60!black}{+1.7} 
& \textcolor{green!60!black}{+3.8} 
& \textcolor{green!60!black}{+5.2} 
& \textcolor{green!60!black}{+0.084} 
& \textcolor{green!60!black}{+0.062} 
& \textcolor{green!60!black}{+0.164} \\

PyramidKV 
& 58.2 & 54.3 & 43.4 & 44.1 & 39.4 & 29.1 & 62.9 & 60.9 & 54.8 & 49.1 & 47.1 & 40.8 & 0.621 & 0.553 & 0.407 \\
+ MixKV 
& 60.8 & 57.2 & 45.1 & \textbf{49.7} & 43.7 & 32.0 & 64.9 & 63.8 & 57.8 & 50.8 & 47.5 & 41.3 & 0.687 & 0.656 & 0.466 \\
\rowcolor{gray!10}
+ ECOKV
& \textbf{61.5} & \textbf{59.0} & \textbf{56.1} & 48.9 & \textbf{46.9} & \textbf{44.1} & \textbf{65.1} & \textbf{63.9} & \textbf{62.0} & \textbf{51.2} & \textbf{50.7} & \textbf{49.3} & \textbf{0.720} & \textbf{0.702} & \textbf{0.662} \\
\rowcolor{gray!20}
$\Delta_{\text{baseline}}$
& \textcolor{green!60!black}{+0.7}
& \textcolor{green!60!black}{+1.8}
& \textcolor{green!60!black}{+11.0}
& \textcolor{red!60!black}{-0.8}
& \textcolor{green!60!black}{+3.2}
& \textcolor{green!60!black}{+12.1}
& \textcolor{green!60!black}{+0.2}
& \textcolor{green!60!black}{+0.1}
& \textcolor{green!60!black}{+4.2}
& \textcolor{green!60!black}{+0.4}
& \textcolor{green!60!black}{+3.2}
& \textcolor{green!60!black}{+8.0}
& \textcolor{green!60!black}{+0.033}
& \textcolor{green!60!black}{+0.046}
& \textcolor{green!60!black}{+0.196} \\

AdaKV 
& 59.6 & 55.9 & 48.7 & 45.1 & 40.4 & 32.8 & 62.9 & 60.5 & 56.9 & 50.4 & 47.8 & 44.6 & 0.646 & 0.566 & 0.440 \\
+ MixKV 
& 61.3 & 58.3 & 50.8 & 49.8 & 44.9 & 36.6 & 65.3 & 63.7 & 59.6 & 50.9 & 48.5 & 45.2 & 0.704 & 0.660 & 0.509 \\
\rowcolor{gray!10}
+ ECOKV
& \textbf{62.3} & \textbf{60.3} & \textbf{56.1} & 49.0 & \textbf{47.3} & \textbf{44.4} & \textbf{65.4} & \textbf{64.5} & \textbf{61.8} & \textbf{51.9} & \textbf{51.1} & \textbf{49.5} & \textbf{0.721} & \textbf{0.711} & \textbf{0.663} \\
\rowcolor{gray!20}
$\Delta_{\text{baseline}}$
& \textcolor{green!60!black}{+1.0}
& \textcolor{green!60!black}{+2.0}
& \textcolor{green!60!black}{+5.3}
& \textcolor{red!60!black}{-0.8}
& \textcolor{green!60!black}{+2.4}
& \textcolor{green!60!black}{+7.8}
& \textcolor{green!60!black}{+0.1}
& \textcolor{green!60!black}{+0.8}
& \textcolor{green!60!black}{+2.2}
& \textcolor{green!60!black}{+1.0}
& \textcolor{green!60!black}{+2.6}
& \textcolor{green!60!black}{+4.3}
& \textcolor{green!60!black}{+0.017}
& \textcolor{green!60!black}{+0.051}
& \textcolor{green!60!black}{+0.154} \\
SparseMM 
& 61.6 & 60.8 & 57.6 & \textbf{51.9} & \textbf{50.7} & 46.2 & 65.1 & 64.7 & 62.8 & 51.9 & 51.2 & 48.9 & 0.680 & 0.634 & 0.524 \\

+ MixKV 
& 61.9 & 61.0 & 59.2 & 50.8 & 50.4 & \textbf{49.5} & 65.2 & \textbf{65.0} & 64.4 & 51.8 & \textbf{51.5} & 50.6 & 0.682 & 0.652 & 0.575 \\
\rowcolor{gray!10}
+ ECOKV
& \textbf{61.9} & \textbf{61.5} & \textbf{61.7} & 51.1 & 50.3 & 48.3 & \textbf{65.2} & 64.9 & \textbf{64.6} & \textbf{51.8} & 51.2 & \textbf{50.6} & \textbf{0.693} & \textbf{0.683} & \textbf{0.667} \\
\rowcolor{gray!20}
$\Delta_{\text{baseline}}$
& \textcolor{green!60!black}{+0.0}
& \textcolor{green!60!black}{+0.5}
& \textcolor{green!60!black}{+2.5}
& \textcolor{green!60!black}{+0.3}
& \textcolor{red!60!black}{-0.1}
& \textcolor{red!60!black}{-1.2}
& \textcolor{green!60!black}{+0.0}
& \textcolor{red!60!black}{-0.1}
& \textcolor{green!60!black}{+0.2}
& \textcolor{green!60!black}{+0.0}
& \textcolor{red!60!black}{-0.3}
& \textcolor{green!60!black}{+0.0}
& \textcolor{green!60!black}{+0.011}
& \textcolor{green!60!black}{+0.031}
& \textcolor{green!60!black}{+0.092} \\
\midrule
FullKV
& \textbf{63.7} & - & - & \textbf{52.7} & - & - & \textbf{65.8} & - & - & \textbf{52.9} & - & - & \textbf{0.702} & - & - \\

\midrule
\multicolumn{16}{c}{\textbf{Qwen2-VL-7B-Instruct}} \\
\midrule
SnapKV 
& 88.0 & 80.1 & 66.5 & 77.3 & 71.9 & 62.4 & 80.3 & 77.5 & 69.9 & 81.3 & 79.6 & 75.5 & 1.360 & 1.142 & 0.794 \\
+ MixKV 
& 90.5 & 82.6 & 67.9 & 79.3 & 75.4 & 66.0 & 81.9 & 80.6 & 72.5 & 81.6 & 81.2 & 77.6 & 1.470 & 1.342 & 0.878 \\
\rowcolor{gray!10}
+ ECOKV
& \textbf{93.4} & \textbf{90.4} & \textbf{81.8} & \textbf{80.9} & \textbf{78.9} & \textbf{74.8} & \textbf{82.1} & \textbf{82.1} & \textbf{78.8} & \textbf{81.8} & \textbf{81.6} & \textbf{81.0} & \textbf{1.517} & \textbf{1.496} & \textbf{1.294} \\
\rowcolor{gray!20}
$\Delta_{\text{baseline}}$
& \textcolor{green!60!black}{+2.9} & \textcolor{green!60!black}{+7.8} & \textcolor{green!60!black}{+13.9} & \textcolor{green!60!black}{+1.6} & \textcolor{green!60!black}{+3.5} & \textcolor{green!60!black}{+8.8} & \textcolor{green!60!black}{+0.2} & \textcolor{green!60!black}{+1.5} & \textcolor{green!60!black}{+6.3} & \textcolor{green!60!black}{+0.2} & \textcolor{green!60!black}{+0.4} & \textcolor{green!60!black}{+3.4} & \textcolor{green!60!black}{+0.047} & \textcolor{green!60!black}{+0.154} & \textcolor{green!60!black}{+0.416} \\
\color{black}

PyramidKV 
& 81.7 & 74.0 & 59.9 & 74.5 & 67.9 & 56.8 & 78.3 & 74.6 & 65.3 & 81.1 & 79.2 & 73.5 & 1.115 & 0.951 & 0.569 \\
+ MixKV 
& 84.0 & 76.3 & 60.8 & 76.6 & 72.6 & 58.4 & 80.4 & 77.1 & 67.0 & 81.3 & 80.7 & 75.5 & 1.348 & 1.119 & 0.633 \\
\rowcolor{gray!10}
+ ECOKV
& \textbf{89.0} & \textbf{82.4} & \textbf{71.3} & \textbf{78.5} & \textbf{75.7} & \textbf{63.7} & \textbf{81.4} & \textbf{78.2} & \textbf{71.6} & \textbf{81.9} & \textbf{81.3} & \textbf{79.5} & \textbf{1.426} & \textbf{1.250} & \textbf{0.913} \\
\rowcolor{gray!20}
$\Delta_{\text{baseline}}$
& \textcolor{green!60!black}{+5.0} & \textcolor{green!60!black}{+6.1} & \textcolor{green!60!black}{+10.5} & \textcolor{green!60!black}{+1.9} & \textcolor{green!60!black}{+3.1} & \textcolor{green!60!black}{+5.3} & \textcolor{green!60!black}{+1.0} & \textcolor{green!60!black}{+1.1} & \textcolor{green!60!black}{+4.6} & \textcolor{green!60!black}{+0.6} & \textcolor{green!60!black}{+0.6} & \textcolor{green!60!black}{+4.0} & \textcolor{green!60!black}{+0.078} & \textcolor{green!60!black}{+0.181} & \textcolor{green!60!black}{+0.280} \\

AdaKV 
& 87.4 & 81.2 & 67.1 & 77.8 & 71.0 & 62.1 & 79.9 & 77.0 & 70.3 & 80.8 & 79.6 & 75.9 & 1.345 & 1.146 & 0.775 \\
+ MixKV 
& 90.3 & 82.1 & 67.8 & 79.3 & 74.7 & 65.5 & 81.8 & 79.6 & 71.2 & 81.5 & 80.9 & 77.4 & 1.448 & 1.275 & 0.878 \\
\rowcolor{gray!10}
+ ECOKV
& \textbf{93.1} & \textbf{90.5} & \textbf{82.7} & \textbf{81.1} & \textbf{79.3} & \textbf{74.9} & \textbf{82.2} & \textbf{81.7} & \textbf{78.7} & \textbf{81.8} & \textbf{81.6} & \textbf{80.4} & \textbf{1.490} & \textbf{1.456} & \textbf{1.301} \\
\rowcolor{gray!20}
$\Delta_{\text{baseline}}$
& \textcolor{green!60!black}{+2.8} & \textcolor{green!60!black}{+8.4} & \textcolor{green!60!black}{+14.9} & \textcolor{green!60!black}{+1.8} & \textcolor{green!60!black}{+4.6} & \textcolor{green!60!black}{+9.4} & \textcolor{green!60!black}{+0.4} & \textcolor{green!60!black}{+2.1} & \textcolor{green!60!black}{+7.5} & \textcolor{green!60!black}{+0.3} & \textcolor{green!60!black}{+0.7} & \textcolor{green!60!black}{+3.0} & \textcolor{green!60!black}{+0.042} & \textcolor{green!60!black}{+0.181} & \textcolor{green!60!black}{+0.423} \\

SparseMM 
& 93.5 & 91.5 & 84.9 & 81.2 & 79.0 & 74.3 & 82.0 & 81.6 & 77.3 & 82.0 & 81.5 & 80.1 & 1.482 & 1.430 & 1.038 \\
+ MixKV 
& 93.8 & 92.7 & 86.4 & \textbf{82.0} & 81.0 & 77.1 & 82.0 & \textbf{82.0} & 80.9 & 81.6 & \textbf{81.8} & 81.4 & 1.480 & 1.459 & 1.303 \\
\rowcolor{gray!10}
+ ECOKV
& \textbf{93.9} & \textbf{93.5} & \textbf{93.2} & 81.9 & \textbf{81.4} & \textbf{80.3} & \textbf{82.1} & \textbf{82.0} & \textbf{81.6} & \textbf{82.0} & 81.6 & \textbf{81.6} & \textbf{1.497} & \textbf{1.510} & \textbf{1.494} \\
\rowcolor{gray!20}
$\Delta_{\text{baseline}}$
& \textcolor{green!60!black}{+0.1}
& \textcolor{green!60!black}{+0.8}
& \textcolor{green!60!black}{+6.8}
& \textcolor{red!60!black}{-0.1}
& \textcolor{green!60!black}{+0.4}
& \textcolor{green!60!black}{+3.2}
& \textcolor{green!60!black}{+0.1}
& \textcolor{green!60!black}{+0.0}
& \textcolor{green!60!black}{+0.7}
& \textcolor{green!60!black}{+0.4}
& \textcolor{red!60!black}{-0.2}
& \textcolor{green!60!black}{+0.2}
& \textcolor{green!60!black}{+0.017}
& \textcolor{green!60!black}{+0.051}
& \textcolor{green!60!black}{+0.191} \\
\midrule
FullKV
& \textbf{93.8} & - & - & \textbf{82.3} & - & - & \textbf{82.1} & - & - & \textbf{81.8} & - & - & \textbf{1.477} & - & - \\

\bottomrule
\end{tabular}
}
\label{tab:main_full}
\end{table*}

\section{Experiment}

\subsection{Experiment Setup}

We evaluate our method on two multimodal large language models: LLaVA-NeXT-Mistral-7B~\citep{liu2024llavanext} and Qwen2-VL-7B-Instruct~\citep{wang2024qwen2}. For standard multimodal benchmarks, we assess performance on OCRBench~\citep{liu2024ocrbench}, ChartQA~\citep{masry2022chartqa}, TextCaps~\citep{sidorov2020textcaps}, TextVQA~\citep{singh2019towards}, and DocVQA~\citep{mathew2021docvqa} following the settings in previous works~\citep{wang2025sparsemm, liu2026mixing}. We use LMMs-Eval~\citep{lmms_eval2024} as a unified evaluation framework across all benchmarks. To demonstrate plug-and-play compatibility, we integrate ECOKV with four state-of-the-art KV cache eviction methods: SnapKV~\citep{li2024snapkv}, PyramidKV~\citep{cai2024pyramidkv}, AdaKV~\citep{feng2024ada}, and SparseMM~\citep{wang2025sparsemm}. We compare against MixKV~\citep{liu2026mixing} and full KV cache (FullKV) as baselines. Concise introductions of the baseline methods and how we integrate ECOKV with them are provided in supplementary section \ref{baselines}. The benchmark content and the score metrics reported are listed in supplementary section \ref{benchmark}. The hyperparameters for evaluation are provided in supplementary section \ref{runtime}. All experiments are conducted on NVIDIA GeForce RTX 4090 GPUs with 24GB memory.

\subsection{Main Results}
\subsubsection{Multimodal Question Answering Benchmarks}

As shown in Table \ref{tab:main_full}, our method achieves state-of-the-art performance and can be seamlessly incorporated with baselines. The performance improvements are consistently observed across diverse benchmarks, model architectures, and budget settings. Notably, when the budget is set to 256, our approach already matches or even surpasses the performance of FullKV. Under more constrained budgets, our method significantly outperforms all baselines, demonstrating its effectiveness in retaining critical information. We also report $\Delta_{\text{baseline}}$, which denotes the performance difference between our approach and MixKV, to highlight the advantage of our method over relying solely on a single cosine similarity metric.

\subsubsection{Long-context VQA} \label{long_context}

We further evaluate our model's ability on Long context VQA datasets: MP-DocVQA~\citep{tito2023hierarchical}. MP-DocVQA is a vision question answering dataset that features multiple-page documents as input, which has a longer context length and requires accurate extraction of helpful information. The experiments are conducted on Qwen2-VL-7B-Instruct. As shown in Table \ref{mpdocvqa}, our method consistently outperforms baselines across different compression ratios, validating its effectiveness in long-context settings. 

\begin{table}[t]
\centering
\caption{Evaluation results on MP-DocVQA.} 
\scalebox{0.95}{
    \begin{tabular}{lccc}
    \toprule
    \multirow{2}{*}{\textbf{Methods}}
    & \multicolumn{3}{c}{\bf Budget Size} \\
    \cmidrule(lr){2-4}
    & 256 & 128 & 64\\
    \midrule 
    SnapKV & 55.0 & 50.3 & 42.1 \\
    SnapKV + MixKV & 56.2 & 51.7 & 42.8 \\
    SnapKV + ECOKV & \textbf{58.0} & \textbf{56.1} & \textbf{51.3} \\
    \midrule 
    PyramidKV & 51.4 & 46.4 & 38.1 \\
    PyramidKV + MixKV & 52.6 & 47.8 & 38.6\\
    PyramidKV + ECOKV & \textbf{55.5} & \textbf{51.5} & \textbf{44.5} \\
    \bottomrule
    \end{tabular}
}
\label{mpdocvqa}
\end{table}


\subsection{Ablation Studies}
\label{subsec:ablation}
We conduct ablation studies to verify the effectiveness of the proposed components. The experiments are conducted on Qwen2-VL-7B-Instruct with SnapKV and PyramidKV to demonstrate consistency across different eviction frameworks, with ChartQA selected as the benchmark. 
\begin{figure}[h]
  \centering
  \includegraphics[width=\linewidth]{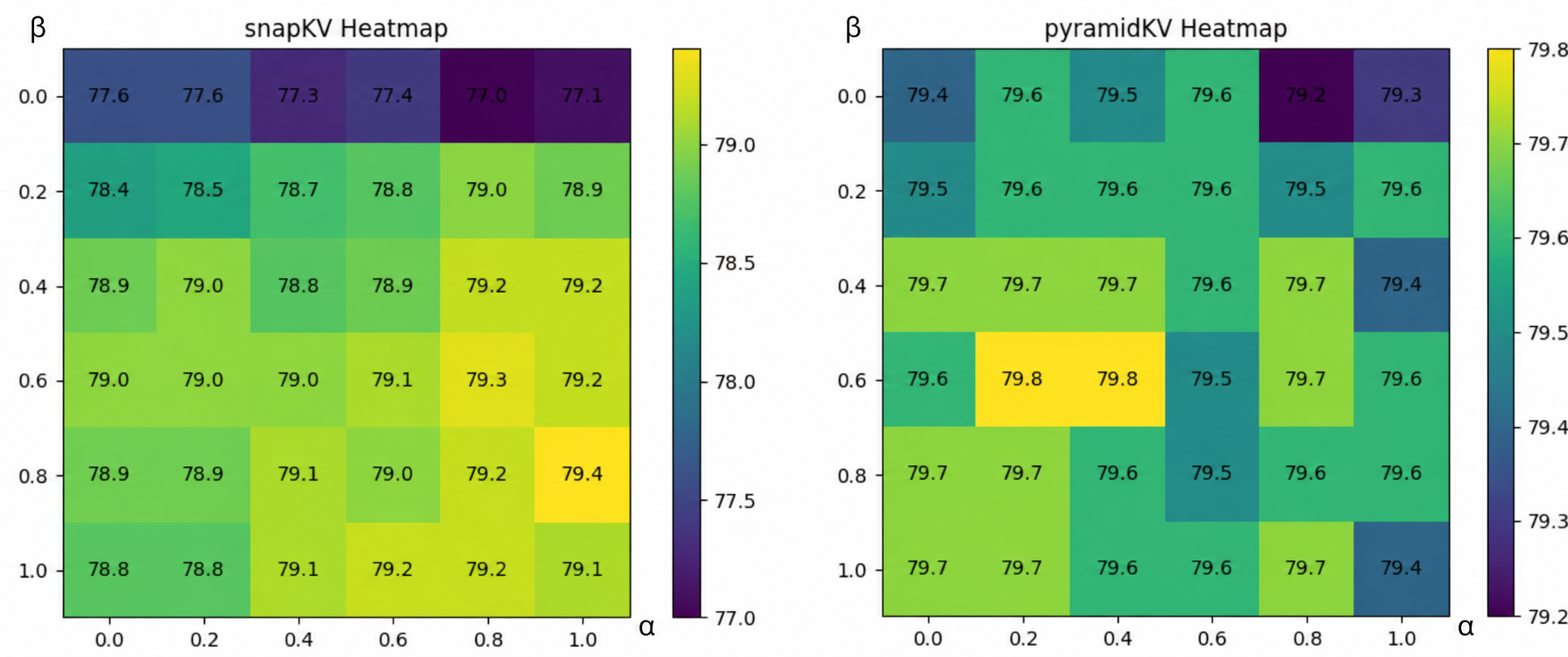}
  \caption{Performance under different combination weights $\alpha$ and $\beta$. Smaller window size is not applied to isolate gains from metric design.}
    \label{figure:alpha_ablation}
\end{figure}
\paragraph{Effect of Combination Weight}
We investigate the weighting coefficients $\alpha$ and $\beta$, which balance Euclidean distance and cosine similarity in the diversity metric and redundancy estimator. As shown in Figure~\ref{figure:alpha_ablation}, $\beta \approx 0.6$ consistently achieves strong performance, while optimal $\alpha$ varies between baseline methods. Notably, the optimal combination outperforms both pure Euclidean distance ($\alpha=\beta=0$) and pure cosine similarity ($\alpha=\beta=1$) configurations, demonstrating their complementary geometric properties. Although the optimal weighting coefficients vary across models and benchmarks, a fixed setting of $\alpha = 0.8$  and $\beta = 0.6$ consistently achieves near-optimal performance, eliminating the need for exhaustive hyperparameter search.

\paragraph{Analysis of Observation Window Size}
We examine the effect of observation window size by evaluating three different window sizes across various model and benchmark settings. As shown in Table~\ref{window_uni}, employing a tiny observation window (window size = 4) or preserving only the most recent token (window size = 1) consistently yields better performance across different benchmarks and model architectures.  This supports our hypothesis that large observation windows consume the available budget, limiting the capacity for diversity- and importance-based selection. Notably, removing the observation window by only preserving and using the last query to assess importance scores still yields comparable or even superior result, indicating that the observation window potentially limit the model's capability. We provide fine-grained window size ablation studies and more analysis in supplementary section \ref{b-1}.




\begin{table}[t]
\centering
\caption{Reducing observation window size yields universal improvement.}
\label{window_uni}

\begin{minipage}[t]{0.48\textwidth}
\centering
\small
\begin{tabular}{lccc}
\toprule
\textbf{Window size} & 32 & 4 & 1 \\
\midrule
OCR     & 68.1 & \textbf{74.8} & 73.6 \\
ChartQA & 79.1 & \textbf{81.0} & 80.9 \\
TextVQA & 74.4 & 78.8 & \textbf{80.2} \\
\bottomrule
\end{tabular}
\\[2pt]
\textbf{(a) Qwen2-VL-7B-Instruct}
\vspace{0.5cm}
\end{minipage}
\hfill
\begin{minipage}[t]{0.48\textwidth}
\centering
\small
\begin{tabular}{lccc}
\toprule
\textbf{Window size} & 32 & 4 & 1 \\
\midrule
OCR     & 36.2 & \textbf{43.7} & 39.7 \\
ChartQA & 43.8 & 48.8 & \textbf{49.0} \\
TextVQA & 60.1 & \textbf{63.0} & 62.8 \\
\bottomrule
\end{tabular}
\\[2pt]
\textbf{(b) LLaVA-NeXT-Mistral-7B}
\end{minipage}

\end{table} 

\paragraph{Normalization Strategy}

To assess the robustness of our min-max normalization to outliers and alternative scaling strategies, we conduct additional ablations on Qwen2-VL-7B-Instruct using ChartQA with SnapKV. Specifically, we compare our original min-max normalization (Eq.~2) with winsorized min-max normalization and per-head MAD scaling (with temperature set to 1). Winsorized min-max normalization reduces the influence of extreme values by clipping outliers, whereas MAD scaling provides a distribution-aware alternative based on the median absolute deviation. As shown in Table \ref{tab:normalization}, all three strategies achieve comparable performance, suggesting that outliers do not have a significant impact on our method.

\begin{table}[t]
\centering
\caption{Comparison of different normalization methods under different cache budgets.}
\label{tab:normalization}
\resizebox{\linewidth}{!}{
\begin{tabular}{lccc}
\toprule
\textbf{Method \textbackslash Budget} & \textbf{256} & \textbf{128} & \textbf{64} \\
\midrule
Original min-max
    & \textbf{81.8} & \textbf{81.6} & \textbf{81.0} \\
1\%-99\% winsorized min-max
    & \textbf{81.8} & \textbf{81.6} & 80.8 \\
5\%-95\% winsorized min-max
    & 81.7 & \textbf{81.6} & 80.8 \\
MAD
    & 81.7 & \textbf{81.6} & 80.9 \\
\bottomrule
\end{tabular}
}
\end{table}


\paragraph{Efficiency Analysis}

To evaluate efficiency, we construct synthetic inputs with a context length of 8192 tokens and measure key metrics, including cache size, peak memory usage, and total execution time for generating 512 output tokens. For each method, experiments are repeated three times, and we report the average results to ensure robustness. All methods are evaluated under identical hardware conditions to guarantee a fair comparison. The budget size is set to 64 for all methods except FullKV. As shown in Table~\ref{tab:efficiency}, incorporating ECOKV achieves efficiency comparable to baseline methods. This is because the additional computational overhead introduced by Euclidean distance computation is minimal, and is partially offset by reduced computation due to the smaller window size. In addition to the synthetic results, we also provide efficiency results on the actual benchmark in supplemental material section \ref{b-3} to further validate ECOKV's efficiency in practical settings.

\begin{table}[t]
\centering
\caption{Average performance over 3 runs.}
\scalebox{0.7}{
\begin{tabular}{lccc}
\toprule
Method & Cache (GB) & Peak Memory (GB) & Total Time (s)\\
\midrule
FullKV & 0.438 & 20.574 & 20.459  \\
\midrule
SnapKV & 0.003 & 20.140  & 16.536 \\
SnapKV + MixKV & 0.003  & 20.140 & 16.483 \\
SnapKV + ECOKV & 0.003  & 20.140  & 16.408  \\
\midrule
PyramidKV & 0.004  & 20.140  & 16.662 \\
PyramidKV + MixKV & 0.004  & 20.140  & 16.550 \\
PyramidKV + ECOKV & 0.004  & 20.140  & 16.540 \\
\bottomrule
\end{tabular}
}
\label{tab:efficiency}
\end{table}

\subsection{Futher Analysis} \label{further}

\paragraph{Correlation Between Importance and Diversity.}

\begin{figure}[t]
  \centering
  \vspace{-0.2cm}\includegraphics[width=0.88\linewidth]{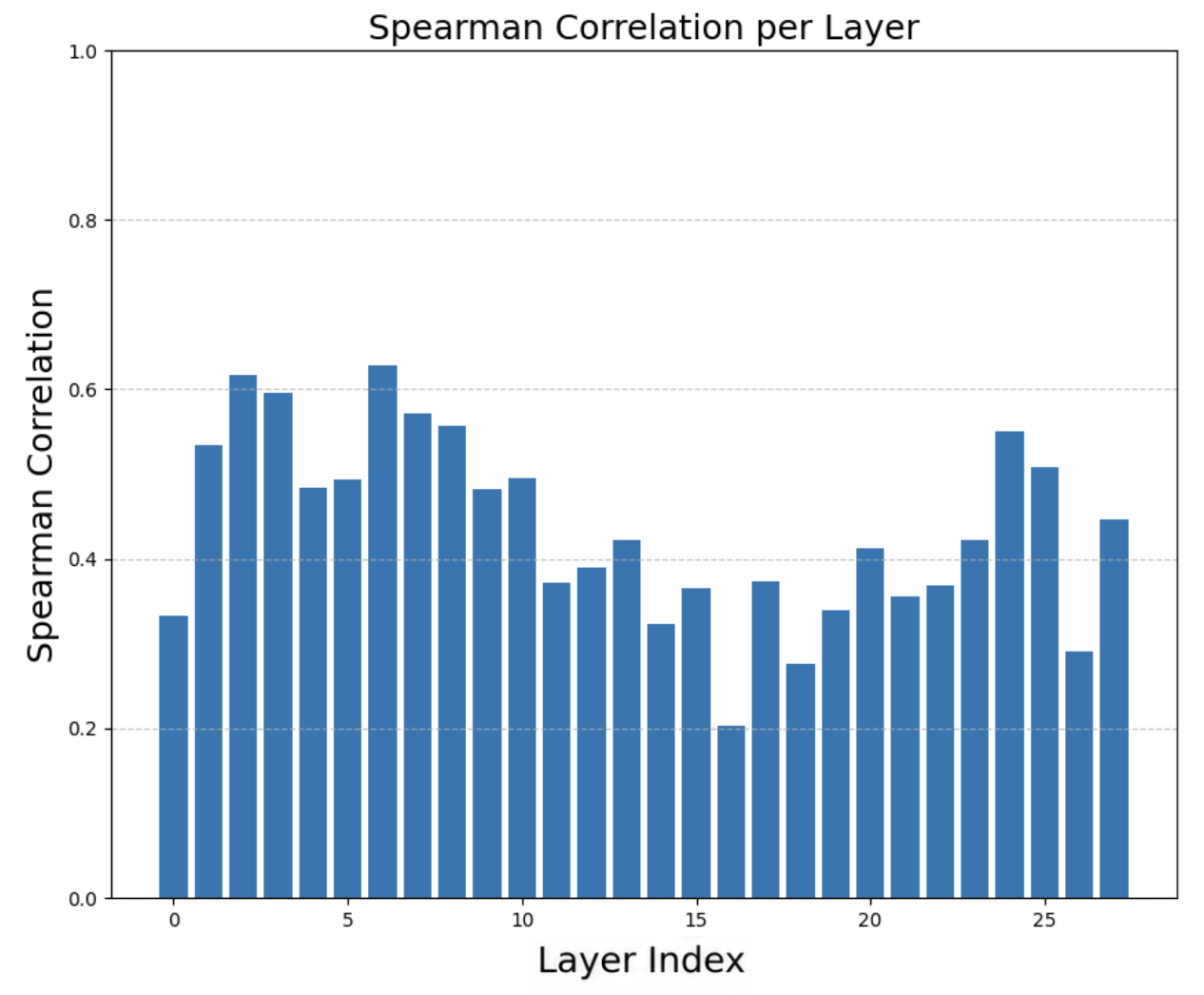}
  \caption{Correlation between importance and diversity metrics.}
  \label{figure:importance_diversity_correlation}
\end{figure}

We analyze the correlation between importance and diversity scores to understand whether they provide complementary or redundant signals for key-value selection. The experiment is conducted on Qwen2-VL-7B-Instruct using 100 randomly sampled test cases from ChartQA. Figure~\ref{figure:importance_diversity_correlation} reports the average Spearman correlation between importance-based and diversity-based rankings across layers, ordered from early to late layers. We observe a generally low to moderate correlation, suggesting that importance and diversity metrics prioritize keys in different orders. In addition, the correlation is lower in intermediate layers, indicating that the diversity metric has the greatest influence on key-value selection in those layers, where relying solely on importance-based scoring would result in a less representative retained set. We provide results on OCRBench, TextVQA, and TextCaps in supplementary section \ref{b-4} and exhibit similar patterns, consistently supporting these findings.\\

\section{Conclusion}

We present ECOKV, a novel KV cache eviction method that integrates complementary diversity metrics to provide a richer geometric characterization of token diversity.  
We employ the proposed composite metric to prioritize and retain critical key–value pairs, estimate layer-wise redundancy, and show that reducing the observation window can further improve performance under tight budget constraints. Experiments demonstrate that ECOKV consistently improves existing eviction methods across various benchmarks and compression ratios. Our analysis also validates the effectiveness of each component and the correlation between importance and diversity across layers.

{
    \small
    \bibliographystyle{ieeenat_fullname}
    \bibliography{main}
}

\clearpage
\appendix
\section{Experimental Setup Details}
\subsection{Baselines} \label{baselines}

We briefly introduce the mechanisms proposed in the baseline methods. We incorporate our ECOKV with these methods by replacing their scoring function $S$ with our approach and reduce the observation window size to 4. \\

\begin{itemize}
    \item \textbf{SnapKV}~\citep{li2024snapkv}:

    SnapKV compresses KV caches by selecting clustered important KV positions for each attention head during the prefilling stage. They introduce the idea of an observation window and design a voting mechanism to determine the top-K important keys:
    \[
    S = \sum_{j=0}^{N} \mathbf{w}_{\text{obs}}[:, j, :], \quad \quad \hat{K} = \operatorname{Top}_k(S, k)
    \]

    The attention weight is calculated for each query in the observation window, and accumulated as the importance score $S$. Then the selected key-value pairs index $\hat{K}$ is determined by top-K selection. They further perform pooling to prevent losing details while evicting less important tokens.\\

    \item \textbf{PyramidKV}~\citep{cai2024pyramidkv}:

    PyramidKV dynamically allocates cache budgets across layers as attention is more diffuse in early layers and concentrated in late layers. They first allocate budgets for the top and bottom layers, which is $b^{m-1}=b^{total} / (\beta \cdot m), b^0=2 \cdot b^{total}/m - b^{m-1}$. For the middle layers, the budgets are calculated as follows:

    \[
    b^l = b^0 - \frac{b^0 - b^{m-1}}{m - 1} \times l
    \]

    $\beta$ is a hyperparameter that determines the shape of the pyramid. \\

    \item \textbf{AdaKV}~\citep{feng2024ada}: 

    AdaKV reallocates budgets for attention heads within the same layer. They first concatenate attention weights across head as $A=Cat(\{A_i\})$, then perform top-K selection in $A$. After selecting the retained weights, they count the number of selected weights for each head i, and set the budget for each head accordingly. Aggregating budget within the same layer allows more dynamic allocation for attention heads.\\
    
    \item \textbf{SparseMM}~\citep{wang2025sparsemm}:

    SparseMM utilizes OCRBench~\citep{liu2024ocrbench} to recognize visual heads, which actively contribute to visual understanding. For each attention head, the total budget $b^l_h$ is computed as follows:

    \[
    b^l_h = N + u + b ^ {score} _ {ij}
    \]

    where $N$ is set to 32, preserving keys within the observation window. The term $u$ denotes a uniformly allocated cache that occupies a small portion of the remaining budget and is evenly distributed across all attention heads. Finally, $b ^ {score} _ {ij}$ determines the allocation of the remaining budget based on visual attention scores estimated from the inference on OCRBench.\\
    
    \item \textbf{MixKV}~\citep{liu2026mixing}\label{mixkv}:

    MixKV combines the existing importance metric and the proposed cosine similarity-based diversity metric. They first derive the similarity between each normalized key and the global average:

    \[
\mathbf{K_N}_i = \frac{\mathbf{K}_{i}}{\|\mathbf{K}_{i}\|}, \quad \mu = \frac{1}{T} \sum_{i=1}^T \mathbf{K_N}_{i}
\]
    \[
    s_C(i) = -\mathbf{K_N}_{i} \cdot \mu
    \]

    $\mathbf{K_N}_i$ denotes the normalized key. The diversity score for each key is computed by the dot product between the normalized key and the global average $\mu$. The author then derives the off-diagonal average similarity to represent the redundancy level of the attention head:

    \[
    r_C = \frac{T^2 |\mu|^2 - T}{T(T-1)}
    \]

    The redundancy level then serves as the weight of diversity metric to combine with the importance metric. Before combination, the diversity score $s_C$ is further scaled to match the magnitude of importance scores.

    \[
    \bar{s_C} = \frac{s_C - min}{max-min}, \quad \tilde{s_C} = \bar{s_C} \cdot \frac{mean(S_{imp})}{mean(\bar{s_C})}
    \]

    Where $S_{imp}$ denotes the importance score. The final $\tilde{s_C}$ is then combined with the importance score.
    
\end{itemize}


\subsection{Benchmarks} \label{benchmark}

Our main experiments include the following benchmarks:\\

\begin{itemize}
    \item \textbf{DocVQA}~\citep{mathew2021docvqa}:
    The benchmark evaluates whether the model recognizes the text in the documents and correctly comprehends the content. We report the Average Normalized Levenshtein Similarity (ANLS) scores of the validation split with 5.35k test cases.\\
    \item \textbf{OCRBench}~\citep{liu2024ocrbench}:
    The benchmark includes 1000 question-answer pairs evaluating the comprehensive ability of MLLMs, including Text Recognition, SceneText-Centric VQA, Document-Oriented VQA, Key Information Extraction, and Handwritten Mathematical Expression Recognition. We report the accuracy evaluated on the whole benchmark.\\
    
    \item \textbf{TextVQA}~\citep{singh2019towards}:
    The benchmark contains questions that require the model to read the image and perform reasoning. We report scores of the validation split with 5k test cases.\\
    
    \item \textbf{ChartQA}~\citep{masry2022chartqa}: 
    The benchmark includes real-world charts and human-authored question-answer pairs that evaluate the models' capability on logical and arithmetic operations. We utilize the test split with 2.5k test cases in total.\\
    
    \item \textbf{TextCaps}~\citep{sidorov2020textcaps}:
    The benchmark requires model to comprehend the text on the images and generate the corresponding captions. We report the CIDEr score on the validation set with 6.5k test cases.\\
    
\end{itemize}

We utilize LMMs-Eval~\citep{lmms_eval2024} to evaluate our models with the above benchmarks.

\subsection{Hyperparameters} \label{runtime}

We provide the hyperparameters used for evaluation in Table~\ref{tab:hyperparameters}. The \textit{ratio} and \textit{mask ratio} parameters are only required by certain baseline methods, such as AdaKV and SparseMM.

\begin{table}[t]
\centering
\caption{Hyperparameters} 
\scalebox{0.85}{
\begin{tabular}{lcc}
\toprule
\textbf{Hyperparameters}       & \textbf{ECOKV} & \textbf{Baselines} \\ 
\midrule
ratio                    & 0.1 & 0.1                \\
mask ratio                    & 0.1 & 0.1                \\
temperature                     & 0.0    & 0.0       \\
batch size & 1  & 1               \\
$\alpha$ & 0.8 & - \\
$\beta$ & 0.6 & - \\
window size & 4 & 32 \\
kernel size & 5 & 5 \\
pooling & average pool & average pool \\
fp16                            & TRUE    & TRUE     \\
\bottomrule
\end{tabular}
}
\label{tab:hyperparameters}
\end{table}

\subsection{Runtime} 

Our main experiments are conducted using LMMs-Eval~\citep{lmms_eval2024} without sampling, ensuring deterministic results across repeated runs. We report the approximate runtime for each benchmark in GPU-hours under our experimental configuration in Table~\ref{tab:runtime}. The runtime is measured on machines equipped with RTX 4090 GPUs and may vary depending on the hardware and system workload during reproduction.

\begin{table}[htbp]
\centering
\caption{Runtime in GPU-hours} 
\scalebox{0.7}{
\begin{tabular}{lcc}
\toprule
\textbf{Benchmark}       & \textbf{LLaVA Runtime (GPU-hours)} & \textbf{Qwen Runtime (GPU-hours)} \\ 
\midrule
DocVQA                    & 1.42       & 4.63         \\
OCRBench                    & 0.23      & 0.15          \\
TextVQA                     & 1.22    & 0.88    \\
ChartQA                     & 0.28    & 0.16    \\
TextCaps                     & 1.01    & 0.92    \\

\bottomrule
\end{tabular}
}
\label{tab:runtime}
\end{table}
\section{Experimental Results and Analysis}

\subsection{Observation Window Analysis} \label{b-1}

\begin{table}[h]
\centering
\caption{
Performance under different window size.} 
\scalebox{0.78}{
\begin{tabular}{llcccccc}
\toprule
\multirow{2.5}{*}{\textbf{Methods}} & \multirow{2.5}{*}{\textbf{Budgets}}
& \multicolumn{6}{c}{\bf Window size} \\
\cmidrule(lr){3-8}
& & 32 & 16 & 8 & 4 & 2 & 1\\
\midrule
\multirow{3}{*}{\textsc{SnapKV}} 
& 256 & 81.8 & 81.8 & 81.8 & 81.8 & 81.7 & 81.8\\
& 128 & 80.9 & 81.5 & 81.5 & 81.6 & 81.7 & 81.8 \\
& 64  & 79.1 & 80.0 & 80.7 & 81.0 & 81.0 & 80.9 \\
\midrule
\multirow{3}{*}{\textsc{PyramidKV}}
& 256 & 81.4 & 81.7 & 81.8 & 81.9 & 81.7 & 81.6 \\
& 128 & 80.8 & 81.4 & 81.3 & 81.3 & 81.7 & 81.6 \\
& 64  & 77.8 & 79.2 & 78.9 & 79.5 & 80.7 & 81.2 \\
\bottomrule
\end{tabular}
}
\label{table:window_ablation}
\end{table}

Table \ref{table:window_ablation} presents finer-grained ablation studies on observation window size. The result further reveals that reducing the window size consistently improves performance under tight budget constraints (budget $= 64$ and $128$).

To better understand this phenomenon, we analyze the selection frequency of tokens within the last 32-token region. Specifically, for each token position, we compute the frequency with which it is retained in the KV cache and average the statistics across all layers and attention heads. The experiment is conducted on 4 benchmarks, with 100 randomly sampled test cases. The budget size is set to 64.

\begin{figure*}[t]
  \centering
  \includegraphics[width=\linewidth]{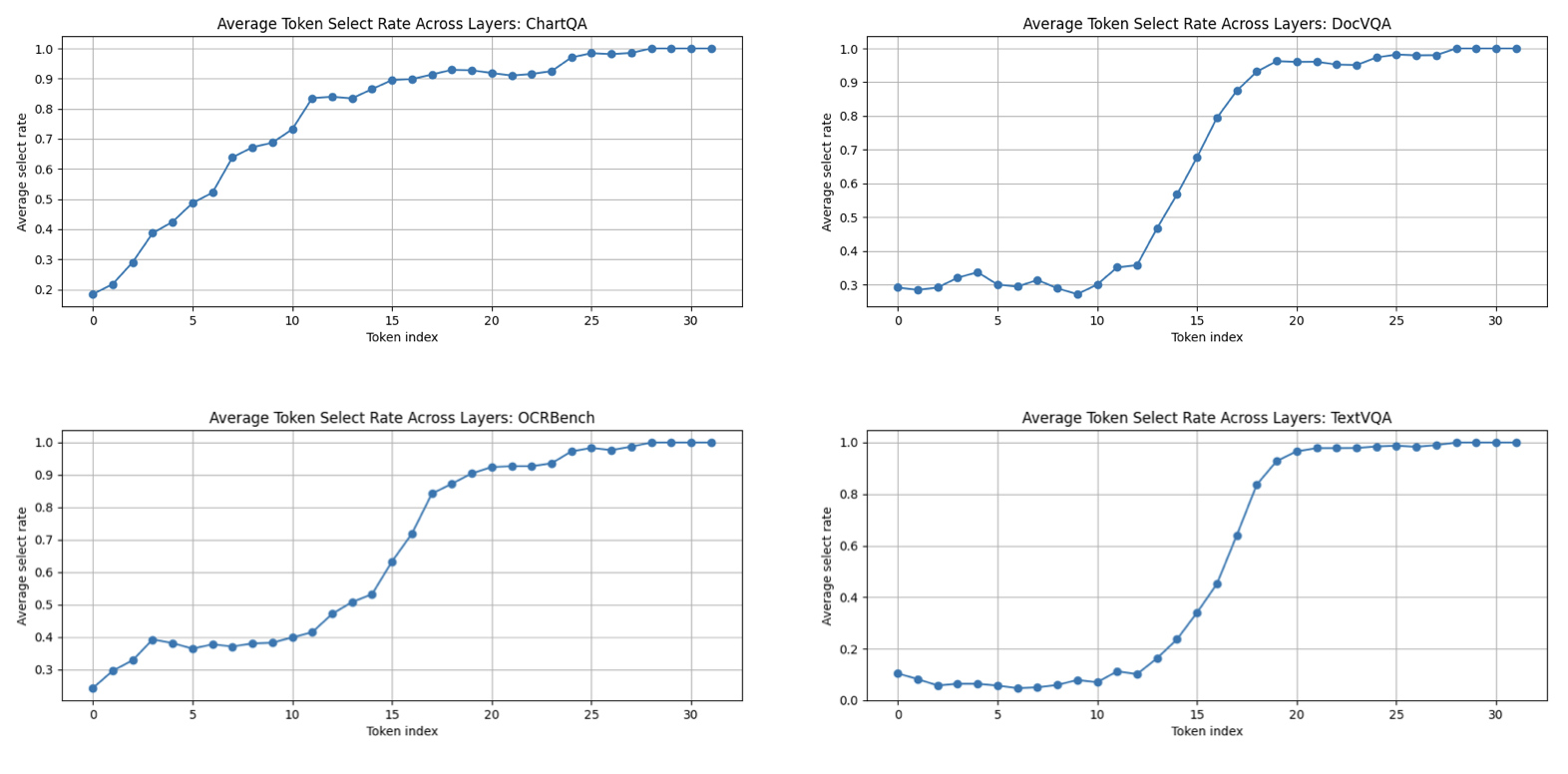}
  \caption{The average selection rate of last 32 tokens across benchmark.}
    \label{figure:window_rate}
\end{figure*}

As illustrated in Figure \ref{figure:window_rate}, different benchmarks exhibit distinct token selection patterns, suggesting that the importance of recent tokens varies across tasks and modalities. Furthermore, we observe that KV cache eviction algorithms naturally preserve recent tokens even without explicitly reserving a large observation window. This finding indicates that enforcing a large fixed observation window is often unnecessary and can instead reduce the budget available for dynamically selected informative tokens.

\subsection{Baselines with Tiny Observation Windows} \label{b-2}

We provide additional experiments by reducing the observation window size for the baseline methods. The experiments are conducted on Qwen2-VL and evaluated on OCRBench. The results in Table \ref{tab:budget} show that smaller observation windows consistently improve baseline performance, confirming that this design choice is effective beyond ECOKV. Moreover, under the same observation window size, ECOKV still consistently outperforms the baselines, demonstrating that our gains cannot be attributed solely to a smaller observation window.

\begin{table}[t]
\centering
\caption{Comparison of different KV cache eviction methods under different cache budgets.}
\label{tab:budget}
\resizebox{\linewidth}{!}{
\begin{tabular}{lccc}
\toprule
\textbf{Method \textbackslash Budget} & \textbf{256} & \textbf{128} & \textbf{64} \\
\midrule
SnapKV (32)
    & 77.3 & 71.9 & 62.4 \\
SnapKV (4)
    & 79.3 & 77.5 & 70.4 \\
SnapKV + MixKV (32)
    & 79.3 & 75.4 & 66.0 \\
SnapKV + MixKV (4)
    & \textbf{81.1} & 78.3 & 74.0 \\
SnapKV + ECOKV
    & 80.9 & \textbf{78.9} & \textbf{74.8} \\
\midrule
PyramidKV (32)
    & 74.5 & 67.9 & 56.8 \\
PyramidKV (4)
    & 76.8 & 72.0 & 54.6 \\
PyramidKV + MixKV (32)
    & 76.6 & 72.6 & 58.4 \\
PyramidKV + MixKV (4)
    & 77.6 & 74.8 & 59.5 \\
PyramidKV + ECOKV
    & \textbf{78.5} & \textbf{75.7} & \textbf{63.7} \\
\bottomrule
\end{tabular}
}
\end{table}

\subsection{Efficiency analysis on real benchmark} \label{b-3}

We replaced the synthetic data with TextVQA and evaluated the efficiency in over 100 randomly sampled test cases. As shown in Table \ref{tab:efficiency_real}, ECOKV achieves efficiency comparable to baselines while requiring a smaller KV cache and lower peak memory than FullKV. The reduction in total inference time is modest because the benchmark produces relatively short outputs, resulting in limited decoding time.

\begin{table}[t]
\centering
\caption{Comparison of cache size, peak memory usage, and total inference time.}
\label{tab:efficiency_real}
\resizebox{\linewidth}{!}{
\begin{tabular}{lccc}
\toprule
\textbf{Method} & \textbf{Cache (GB)} & \textbf{Peak Memory (GB)} & \textbf{Total Time (s)} \\
\midrule
FullKV
    & 0.0280
    & 15.8299
    & 1.0170 \\
SnapKV
    & 0.0034
    & 15.8053
    & 0.9464 \\
SnapKV + MixKV
    & 0.0034
    & 15.8053
    & 1.0543 \\
SnapKV + ECOKV
    & 0.0034
    & 15.8053
    & 1.0113 \\
PyramidKV
    & 0.0036
    & 15.8054
    & 0.9905 \\
PyramidKV + MixKV
    & 0.0036
    & 15.8055
    & 0.9928 \\
PyramidKV + ECOKV
    & 0.0036
    & 15.8054
    & 1.0392 \\
\bottomrule
\end{tabular}
}
\end{table}

\subsection{Correlation between Importance and Diversity} \label{b-4}
\label{more_corr}

\begin{figure*}[t]
  \centering
  \includegraphics[width=\linewidth]{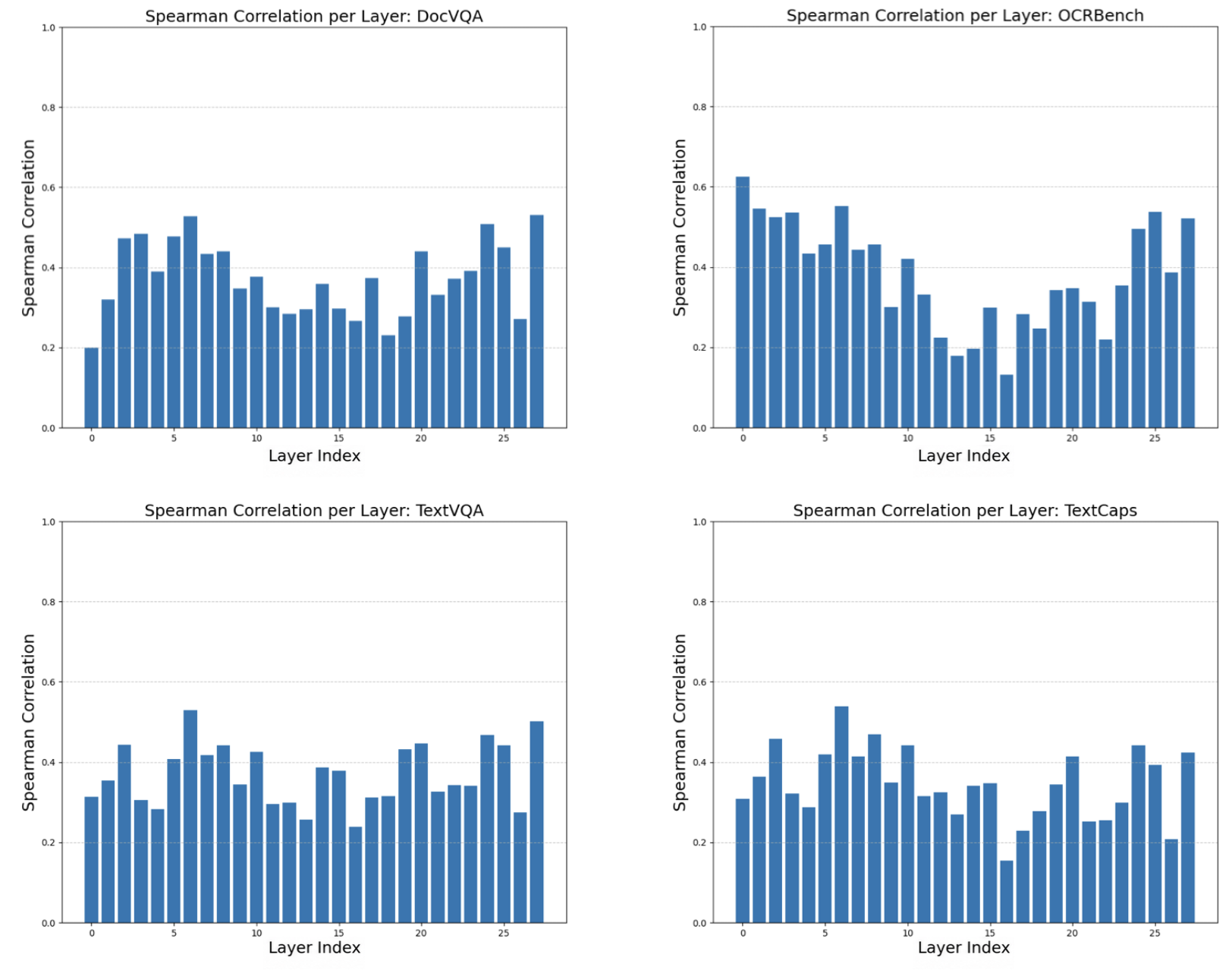}
  \caption{Correlation between importance and diversity across DocVQA, OCRBench, TextVQA, and TextCaps.}
    \label{figure:more_corr}
\end{figure*}

Figure~\ref{figure:more_corr} provides additional correlation results between importance and diversity across various benchmarks, offering further evidence for the analysis presented in the main paper section \ref{further}.

\subsection{Analysis of Diversity Weight}\label{weight_anal}
\begin{figure*}[t!]
  \centering
  \includegraphics[width=\linewidth]{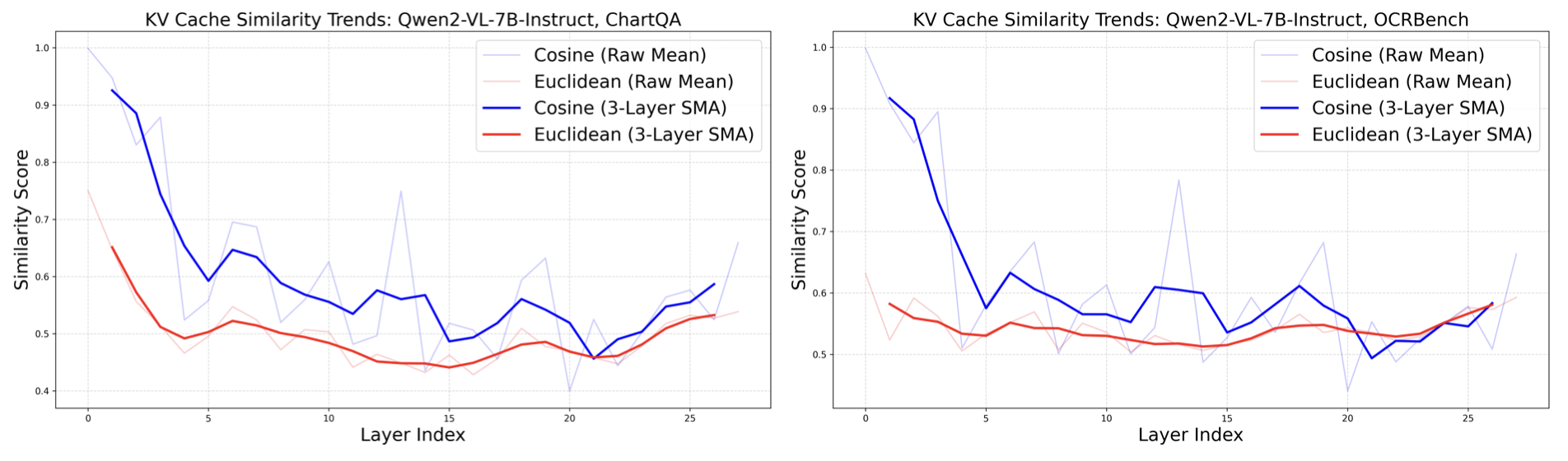}
  \caption{Diversity weight analysis: \color{blue}{cosine similarity} \color{black} vs \color{red}{Euclidean distance} \color{black} on Qwen2-VL-7B-Instruct.}
    \label{figure:diversity_weight}
\end{figure*}

\begin{figure*}[t!]
  \centering
  \includegraphics[width=\linewidth]{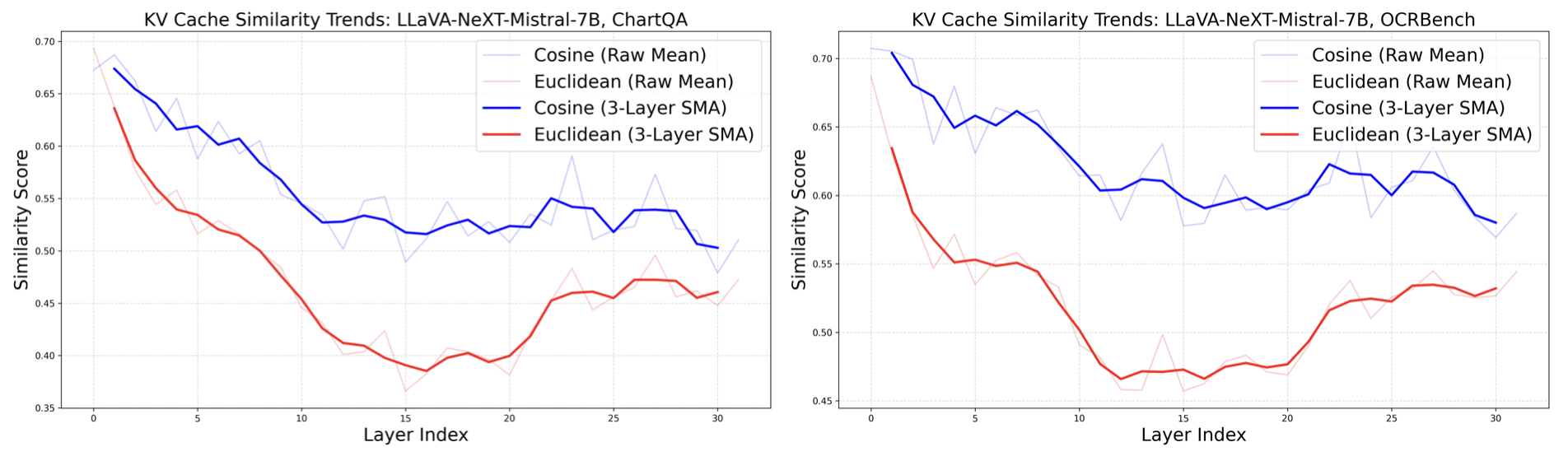}
  \caption{Diversity weight analysis: \color{blue}{cosine similarity} \color{black} vs \color{red}{Euclidean distance} \color{black} on LLaVA-NEXT-Mistral-7B.}
    \label{figure:diversity_weight2}
\end{figure*}

We visualize the diversity weight across layers by sampling 100 test cases from both OCRBench and ChartQA, and compute the average diversity weight. As shown in Figure~\ref{figure:diversity_weight} and Figure~\ref{figure:diversity_weight2}, we plot the layer-averaged weights under cosine similarity and ECOKV Euclidean distance-based metric ($r_C$ and $r_E$ from main paper section \ref{subsec:EDHR}), smoothed with a 3-layer simple moving average (SMA). The two metrics exhibit a similar overall trend across different benchmarks while differing in finer details, suggesting general consistency alongside layer-specific variations. These results also show that redundancy is relatively high in early layers, low in intermediate layers, and gradually increases toward the final layers. \\

\subsection{Global Average Calculation Alternative}

An alternative of computing the global average in Algorithm \ref{alg:similarity} Step 5 is to only average on non-window tokens. To compare with this alternative, we provide additional ablation studies on Qwen-2-VL, OCRBench: ( © indicates computing the centroid only on non-window tokens). The results in Table \ref{exclude} show that the two variants exhibit similar behavior across settings, while our current implementation achieves slightly better performance.

\begin{table}[t]
\centering
\caption{Comparison of alternative global average implementation.}
\begin{tabular}{l|ccc}
\toprule
Method $\backslash$ Budget & 256 & 128 & 64 \\
\hline
SnapKV + ECOKV & 80.9 & 78.9 & 74.8 \\
SnapKV + ECOKV \textcopyright & 80.9 & 78.8 & 74.8 \\
PyramidKV + ECOKV & 78.5 & 75.7 & 63.7 \\
PyramidKV + ECOKV \textcopyright & 78.5 & 75.8 & 62.8 \\
\bottomrule
\end{tabular}
\label{exclude}
\end{table}

\section{Algorithm} \label{alg}

We present the overall algorithm for computing the similarity and redundancy scores in Algorithm \ref{alg:similarity}. The output \textit{Similarity} $S$ is a sequence of diversity scores corresponding to the input keys, while the redundancy $\lambda$ is a scalar that quantifies the overall redundancy of the attention head.

{\footnotesize
\begin{algorithm}[t]
\caption{Similarity and Redundancy Computation}
\label{alg:similarity}
\begin{algorithmic}[1]
\REQUIRE $\mathbf{K} \in \mathbb{R}^{B \times H \times T \times D}$, observation window size $N$
\ENSURE Similarity $S$, redundancy $\lambda$

\STATE $T_v \leftarrow T - N$, \quad $\mathbf{K}_v \leftarrow \mathbf{K}_{:T_v}$

\textbf{Cosine:}
\STATE $\mathbf{K_N} \leftarrow \text{Normalize}(\mathbf{K})$, \quad 
$\boldsymbol{\mu} \leftarrow \frac{1}{T}\sum_t \mathbf{K_N}_t$
\STATE $s_C \leftarrow (\hat{\mathbf{K}}_v \boldsymbol{\mu}^\top + 1)/2$
\STATE $r_C \leftarrow \frac{T\|\boldsymbol{\mu}\|^2 - 1}{T - 1}$

\textbf{Euclidean:}
\STATE $\bar{\mathbf{k}} \leftarrow \frac{1}{T}\sum_i \mathbf{K}_i$, \quad 
$\mathbf{d_E} \leftarrow \|\mathbf{K}_v - \bar{\mathbf{k}}\|_2$
\STATE $s_E \leftarrow  \frac{\mathbf{D} - \min}{\max - \min}$
\STATE $r_E \leftarrow 2 - 2\,\sigma\!\left(\frac{D^h_E}{\tau }\right)$

\textbf{Combine:}
\STATE $S \leftarrow \alpha s_E + (1-\alpha)s_C$
\STATE $\lambda \leftarrow \beta r_E + (1-\beta)r_C$

\RETURN $S, \lambda$
\end{algorithmic}
\end{algorithm}
}
\section{Head-adaptive Redundancy Estimation Alternatives}
\label{alter}
Beyond combining the Euclidean distance with cosine similarity, we design additional heuristics and evaluate their comparative performance.

\subsection{Standard Deviation-based Approach}

As standard deviation primarily estimates the average distance to the mean, which is comparable to our Euclidean implementation in main paper section \ref{subsec:EDHR}. The implementation we use in the main approach can be derived as:

\[
    D_E = \frac{1}{T} \sum_{i=1}^{T} \left\lVert k_i - \bar{k} \right\rVert_2 =  \frac{1}{T} \sum_{i=1}^{T}  \sqrt{(k_i - \bar{k})^2}
\]

where $\bar{k}$ is the mean of keys. We can also compute standard deviation as:

\[
    D_{std}  = \sqrt{ \frac{1}{T} \sum_{i=1}^{T} (k_i - \bar{k})^2}
\]

To compare the two methods, we calculate the standard deviation of the keys and perform scaling as follows:

\begin{equation}
    r_{std} = 1 -  \sigma\!\left(\frac{D_{std}}{\tau}\right)
\end{equation}

where $\tau$ is a temperature parameter. A larger standard deviation indicates a more dispersed distribution, corresponding to higher diversity and lower redundancy. Therefore, we invert the scaled value to obtain a redundancy estimate. $r_{std}$ is then used as the diversity weight.

 We conduct an ablation study by replacing the Euclidean distance with standard deviation and evaluate the resulting performance on ChartQA. As shown in Table \ref{table:std}, the standard-deviation-based approach achieves comparable performance overall, with a slight performance degradation at a budget size of 64.

\begin{table}[t]
\centering
\caption{Standard deviation comparison} 
\begin{tabular}{lccc}
\toprule
\multirow{2}{*}{\textbf{Methods}}
& \multicolumn{3}{c}{\bf Window size} \\
\cmidrule(lr){2-4}
& 256 & 128 & 64\\
\midrule 
Mean Euclidean distance & 81.8 & 81.6 & 81.0 \\
Standard deviation & 81.8 & 81.6 & 80.8 \\
\bottomrule
\end{tabular}
\label{table:std}
\end{table}

\newpage

\end{document}